\documentclass[runningheads]{llncs}
\usepackage{graphicx}
\usepackage{algorithm}
\usepackage{algpseudocode}
\usepackage{float}
\usepackage{caption}
\usepackage{subcaption}
\usepackage{courier}

\begin{document}
\title{Towards Quantifying The Privacy Of Redacted Text}
%
%
\author{Vaibhav Gusain\inst{1}\orcidID{0000-0002-7008-5201} 
\and
Douglas Leith\inst{1}\orcidID{0000-0003-4056-4014} 
\thanks{This work was supported by Science Foundation Ireland grant 16/IA/4610.}
%
\authorrunning{V. Gusain et al.}
%
\institute{Trinity College Dublin, Ireland}\\
} 
\maketitle              
\begin{abstract}
 In this paper we propose use of a k-anonymity-like approach for evaluating the privacy of redacted text.  Given a piece of redacted text we use a state of the art transformer-based deep learning network to reconstruct the original text.   This generates multiple full texts that are consistent with the redacted text, i.e. which are grammatical, have the same non-redacted words etc, and represents each of these using an embedding vector that captures sentence similarity.   In this way we can estimate the number, diversity and quality of full text consistent with the redacted text and so evaluate privacy.

\end{abstract}
\section{Introduction}

Redacting a piece of text involves replacing selected words with an uninformative mask symbol.   Redaction is widely used, but is generally carried out manually and there has been little analysis of the degree of privacy obtained.  Note that evaluating text privacy is generally not straightforward since even when a word is redacted it might still be possible to reliably estimate it from the surrounding text i.e. the context of the redacted word may be revealing.  

Machine learning models for text embedding are often trained by masking out individual words in a piece of text and selecting a model that best reconstructs the missing text.  The idea here is that similar words appear in a similar context.  In particular, transformer-based neural networks such as BART~\cite{lewis2019bart} adopt this approach and achieve state of the performance in many natural language processing tasks.  

Given a piece of redacted text, in this paper we apply transformer-based neural networks to try to reconstruct the original text.  For example, when the text \texttt{he was stationed at singapore} is redacted to \texttt{he was stationed at <mask>} then the top 5 reconstructed text predictions by BART are shown in Table \ref{tab:ex1}.   It can be seen that the reconstructed text is grammatical, consistent with the redacted text (has the same non-redacted words etc) and plausible even though in this example it does not correctly predict the missing word.

\begin{table}[]
    \small
    \centering
    \begin{tabular}{|l| c |}
    \hline
    he was stationed at $<$mask$>$ &\\
\hline
he was stationed at the & 0.62\\
he was stationed at: & 0.58\\
he was stationed at Gettysburg & 0.49\\
he was stationed at Ft. & 0.48\\
he was stationed at Knox & 0.47\\
\hline
    \end{tabular}
    \quad
    \begin{tabular}{|p{2.75cm}| p{3.5cm} |}
    \hline
    Redacted sentence & BART top prediction\\
    \hline
    $<$mask$>$ was $<$mask$>$ at singapore & This article was originally published at singapore\\
    \hline
    $<$mask$>$ $<$mask$>$ $<$mask$>$ at singapore & A look at singapore \\
    \hline
    $<$mask$>$ $<$mask$>$ $<$mask$>$ $<$mask$>$ singapore& Singapore singapore\\
    \hline
    \end{tabular}
    
    \caption{Left-hand table: Top 5 reconstructions by BART for the redacted sentence \texttt{he was stationed at <mask>}.  The values shown in the second column are the corresponding confidence values output by BART.  Right-hand table: top prediction by BART as the number of redacted words is increased.}
    \label{tab:ex1}
\end{table}

In this paper we study using such predicted reconstructions as the basis for a quantitative privacy metric for redacted text.  This is motivated by the observation that the number of reconstructions that are estimated with high confidence can be expected to provide an approximate k-anonymity~\cite{kann} measure i.e a measure of "Hiding in the crowd" privacy since there are at-least K sentences that are plausibly consistent with the redacted text.   Since the reconstructions are represented as embedded vectors that capture sentence similarity (similar sentences are represented by nearby vectors) then we can also estimate the diversity of the reconstructions.

This work reported here is just a first, exploratory step but we find that this general approach shows promise.

Rather than evaluating k-anonymity and text diversity, we begin by considering the text quality of the predictions since this turns out to be a useful predictor of privacy in coarse classification tasks such as sentiment analysis, news article categorisation and medical condition (e.g. has cancer or not).  We find that there is a thresholding effect, whereby beyond a certain level of redaction the quality tends to drop sharply.   By carrying out simulated attacks against the redacted text we  find that the drop in BART prediction quality strongly correlates with a decrease in attack effectiveness.   The proposed approach therefore has the potential to provide a practical, useful estimate of redacted text privacy.


\subsection{Related Work}\label{relatedwork}
\textit{Text Redaction}. Despite the widespread use of redaction, there has been very little work on quantifying the privacy of redacted text or on evaluating robustness to attacks that seek to generate privacy leaks.  Instead most work to date has focused on identifying personal data with text so that it can be redacted.  See, for example,~\cite{detectstudent} which considers discovery of names, home towns etc in student discussion boards, and also the references therein.  The closest work to the present paper is probably~\cite{privacytransformer} which considers randomly redacting words to ensure a form of differential privacy and evaluates utility using a transformer neural net.  However, there is no evaluation of the robustness of the redacted text to adversarial attacks (which is primarily what we use transformer neural nets for here) and the interpretation of differential privacy in the context of redacted text remains unclear (in \cite{privacytransformer} the surrounding context of a redacted word is ignored, yet will often have an important impact on the degree of privacy achieved).



\textit{Text Reconstruction.}  Predicting missing text has been the subject of a great deal work in recent years.  The state of the art uses transformer-based neural net architectures, following the breakthrough performance achieved by BERT.   BART~\cite{lewis2019bart} is a transformer-based neural net that targets reconstruction of text damaged by spelling mistakes, missing words etc.  Roughly speaking it is an amalgamation of BERT and GPT2, consisting of a bidirectional encoder which is very similar to BERT and a left-to-right decoder which is very similar to GPT2. This design allows BART to even predict arbitrary length of text for a single mask token which cant be achieved with BERT.

\section{Quantifying BART Text Quality}
The right-hand table in Table \ref{tab:ex1} shows how the top predicted sentence reconstruction by BART varies as the number of redacted words is increased.  It can be seen that by the time four out the five words in the sentence are redacted the BART prediction degrades and is no longer grammatical.  In our experiments (see below), we find that this behaviour is a common feature of the BART reconstructions.  Of course, it is quite reasonable behaviour since at this point there is so little information left in the redacted sentence that BART has few clues as to how it might be reconstructed.  Equally, the point where this information loss occurs is obviously also of great interest from a privacy viewpoint. 

Rather than considering just the top prediction by BART, we proceed by considering the top N predictions, typically with N=100.  We then estimate the fraction of these predictions which are not grammatical, and investigate the use of this as a measure of privacy.

In general, it is not trivial to estimate whether a sentence is grammatical or not.  Fortunately we do not need to solve the general problem but can instead exploit the fact that BART predictions tend to either be fairly grammatical or else are grossly non-grammatical e.g. with many repetitions of the same word (as can be seen in Table \ref{tab:ex1}) and/or with many repetitions of punctuation and spurious characters.  That is, the BART predictions tend to either be reasonable text or to be ``gibberish''.

\begin{algorithm}
\caption{Algorithm used to classify BART predictions as gibberish or not. Si is the actual input sentence without the mask, Sp is the BART prediction and C is a hyperparameter that checks the number of overlapping words between Si and Sp. It returns True if the prediction is estimated to be gibberish else it returns false}\label{alg:gibberishCalc}
\begin{algorithmic}
\State gibberish = use Nostrill to check if the Sp is gibberish or not.
\If{gibberish}
    \State return gibberish
\Else
    \State return customGibberish(Si,Sp,C)
\EndIf
\end{algorithmic}
\begin{algorithmic}
\State \textbf{customGibberish($Si,Sp,C$)} :
\State $Si \leftarrow$ number of uniquewords in $Si$ 
\State $Sp \leftarrow$ number of uniquewords in $Sp$
\State $p \leftarrow$ number of common words in $Si$ and $Sp$ / length($Si$)
\State gibber = $p$*100/ length($Si$)
\If{gibber $<=$ $C$}
    \State return True
\Else
    \State return False
\EndIf
\end{algorithmic}
\end{algorithm}

To classify a sentence as gibberish or not, in our experiments we use Algorithm~\ref{alg:gibberishCalc} although other choices are of course possible.   Algorithm~\ref{alg:gibberishCalc} combines a standard gibberish detector Nostrill~\cite{nostrill} with a measure of the fraction of words from the original (non-redacted) sentence that overlap with the predicted sentence.  Hyperparameter $C$ controls the weight attached to each measure.



\section{Experimental Measurements}
\subsection{Datasets Used}
We evaluated performance on five datasets: four standard text classification datasets BBCnews~\cite{bbcnews}, Amazon-Fine-food~\cite{amazonfinefood}, AGnews~\cite{abcnews}, IMDB~\cite{imdb} plus the Medal medical dataset~\cite{medal}.  BBCnews has fives classes (Business, Entertainment, Politics, Sport, Tech), Amazon-Fine-food has review stars and reviews with greater than 3 stars were assigned to one class and the rest to another class, AGnews has four classes (World, Sports, Business, Sci/Tech), IMDB has two classes (positive and negative sentiment), Medal has two classes (text specifically about cancer diseases, plus the rest).  Each dataset was split 80:20 into a training and a test dataset, with the training dataset being available to the adversary but not the test dataset.     The datasets are sampled so that they are balanced by category.

\subsection{Threat Model}
The attacker can observe redacted text, and a training data subset of each dataset.  The redacted text is derived from held out data not available to the attacker.  The aim of the attacker is to discover the category of the text e.g. for a movie review to discover the sentiment, for a news article to discover the news category.

\subsection{Reconstruction Quality Metric}
For each redacted sentence we take the top 100 reconstruction predictions from BART and apply Algorithm~\ref{alg:gibberishCalc} to classify them as either gibberish or not, assigning value +1 for gibberish and 0 otherwise.  We calculate the mean of these 100 values.

\subsection{Privacy Attack Performance Metric}
Using the training data for each dataset the adversary trains a classifier based on a TFIDF~\cite{tfidf} vectoriser and a logistic regression model (for these datasets it is known that classifiers of this sort are able to achieve high accuracy).  Given redacted text, the attacker then uses this classifier to estimate the category of the text.   We evaluate the success of this reconstruction using the mean accuracy of these predictions i.e. the fraction of redacted sentences for which the category is correctly estimated.   The test data is balanced, so accuracy is an informative performance measure.

\subsection{Redaction Strategy}
For each dataset we encode the words using a TFIDF vectoriser (discarding words with document frequency less than 10\%).  We then vary the level of redaction by replacing a random $X$ percent of words by a mask token, varying $X$ from 0 to 100\%.   Using TFIDF in this way avoid ineffectual masking of stop words and other uninformative words.  Other redaction strategies are, of course, possible.

\subsection{Additional Material}
We will post our implementations and the associated data on github.

\subsection{Results}\label{result}

\begin{figure}[tb]

\centering
     \begin{subfigure}[b]{0.31\textwidth}
         \centering
         {\includegraphics[width=\textwidth]{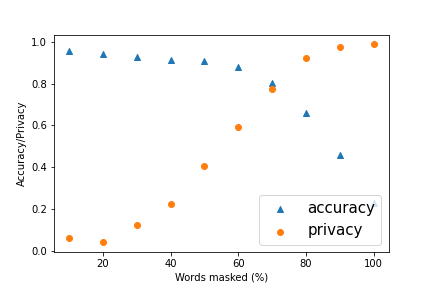}}%
         \caption{BBC}
         \label{bbc}
     \end{subfigure}
     \hfill
     \begin{subfigure}[b]{0.31\textwidth}
         {\includegraphics[width=\textwidth]{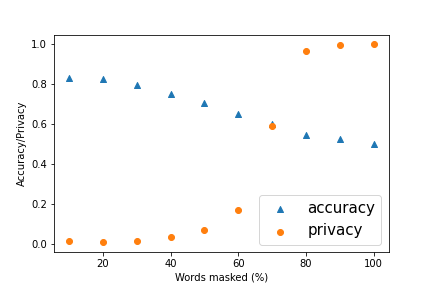}}%
         \caption{Amazon}
         \label{amazonfinefood}
     \end{subfigure}
     \hfill
     \begin{subfigure}[b]{0.31
     \textwidth}
         {\includegraphics[width=\textwidth]{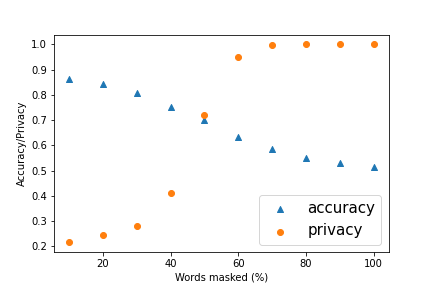}}%
         \hfill 
         \caption{IMDB}
         \label{imdb}
     \end{subfigure}

     \begin{subfigure}[b]{0.31\textwidth}
         {\includegraphics[width=\textwidth]{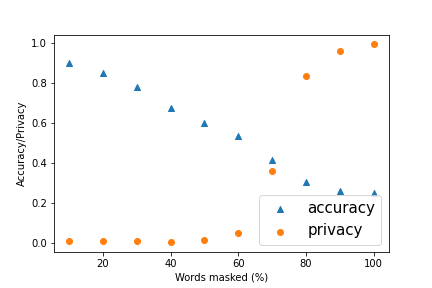}}%
         \caption{Agnews}
         \label{agnews}
     \end{subfigure}
     \hfill
     \begin{subfigure}[b]{0.31\textwidth}
         {\includegraphics[width=\textwidth]{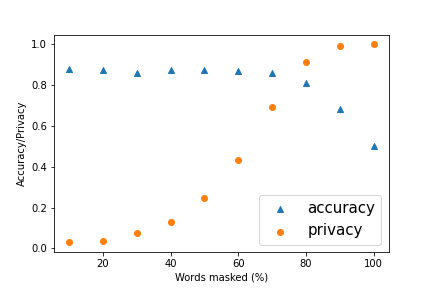}}%
         \caption{Medal}
         \label{medal}
     \end{subfigure}
\caption{Measured privacy metric and attack accuracy for each dataset as the fraction of redacted text is varied from 0 to 100\%.}
\label{resultsGibberish}


\end{figure}

Figure-\ref{resultsGibberish} shows the measured privacy metric and attack accuracy for each dataset as the fraction of redacted text is varied.  It can be seen that as the percentage of masked words is increased the classification accuracy decreases while the privacy metric increases.  

When less than around 20\% of words are redacted, the privacy metric is close to zero for every dataset, indicating that BART consistently reconstructs grammatical sentences that are consistent with the redacted text.  Analysis of the top 100 BART predictions (not included here) show little diversity in the sense that the sentence embedding vectors tend to cluster together.   The attack accuracy is correspondingly also consistently high.  

When greater than around 80\% of words are redacted, then the privacy metric is close to 100\% and the attack accuracy is approximately the reciprocal of the number of categories i.e. comparable with a random coin toss.

Between 20 and 80\% redaction the privacy metric increases and the attack accuracy correspondingly decreases.  By selecting a level of redaction that ensures the privacy metric is above a target threshold, e.g. 70\%, then these measurements indicate that a good level of robustness against the reconstruction attack can be obtained across a wide range of datasets.  

\subsection{Discussion}
Due to lack of space we do not include an evaluation of utility here, which can be expected to degrade as privacy increases.  However, we note briefly that we have evaluated next word prediction performance for the Medal dataset vs privacy and find that the utility remains high even when redaction achieves a high level of resistance against estimation of medical condition.  

We use attack accuracy as a proxy for privacy, since it is difficult to apply standard privacy metrics such as k-anonymity and differential privacy to natural language text data.   However, initial results indicate that it may be possible to estimate a metric similar to k-anonymity by clustering the embedding vectors of the BART predictions and counting the number of distinct clusters.  In the regime where BART predictions are grammatical (redaction level less than 20\% in Figure-\ref{resultsGibberish}) these clusters reflect semantic diversity, whereas in the regime where BART predictions produce lower quality text the clusters tends to become less informative.  However, we leave proper analysis of these aspects to future work.

Initial results also suggest that the nature of the privacy threat is relevant to the level of redaction needed.  To prevent disclosure of broad textual aspects such as sentiment or new category our results show that a high level of redaction is necessary, but preventing disclosure of more fine-grained aspects might be achievable with lower levels of redaction.  Again, we leave further study of this to future work.




%
%
%
\bibliographystyle{splncs04}
\bibliography{sample}

\end{document}